# A New Evaluation Method: Evaluation Data and Metrics for Chinese Grammar Error Correction

Nankai Lin
School of Information Science and Technology
Guangdong University of Foreign Studies
Guangzhou, China
neakail@outlook.com

Yingwen Fu
School of Information Science and Technology
Guangdong University of Foreign Studies
Guangzhou, China
fyinh@foxmail.com

Xiaotian Lin
School of Information Science and Technology
Guangdong University of Foreign Studies
Guangzhou, China
1090030606@qq.com

Ziyu Yang
School of Information Science and Technology
Guangdong University of Foreign Studies
Guangzhou, China
809241889@qq.com

Shengyi Jiang✉
School of Information Science and Technology
Guangdong University of Foreign Studies
Guangzhou, China
jiangshengyi@163.com

## ABSTRACT

As a fundamental task in natural language processing, Chinese Grammatical Error Correction (CGEC) has gradually received widespread attention and become a research hotspot. However, one obvious deficiency for the existing CGEC evaluation system is that the evaluation values are significantly influenced by the Chinese word segmentation results or different language models. The evaluation values of the same error correction model can vary considerably under different word segmentation systems or different language models. However, it is expected that these metrics should be independent of the word segmentation results and language models, as they may lead to a lack of uniqueness and comparability in the evaluation of different methods. To this end, we propose three novel evaluation metrics for CGEC in two dimensions: reference-based and reference-less. In terms of the reference-based metric, we introduce sentence-level accuracy and char-level BLEU to evaluate the corrected sentences. Besides, in terms of the reference-less metric, we adopt char-level meaning preservation to measure the semantic preservation degree of the corrected sentences. We deeply evaluate and analyze the reasonableness and validity of the three proposed metrics, and we expect them to become a new standard for CGEC.

## 1 Introduction

Grammatical Error Correction (GEC) is a fundamental task in natural language processing (NLP) while existing research mainly focuses on English. Since NLPCC 2018 [14] proposed the Chinese Grammatical Error Correction (CGEC) evaluation task, the CGEC task has gradually received widespread attention. However, the test set provided by NLPCC 2018 uses the evaluation metric $F_{0.5}$ from the token level for evaluation, which is based on Chinese word segmentation (CWS) by the pkunlp[1] tool. Notably, "token" in English and Chinese represents different meanings. In English, the token for evaluation is the word, whereas in Latin languages the word is segmented according to spaces so there is no ambiguity in the word segmentation. While in Chinese, the token for evaluation is also the word, but the smallest semantic unit in Chinese is character and a word is made up of one or more characters, so CWS is needed before the evaluation. The evaluation result is to some extent influenced by the CWS algorithms because different CWS algorithms produce different segmentation results, which may result in the following situations:

(1) If the pkunlp tool is no more publicly available, the training and to-be-tested data are split using other CWS tools while the standard test set is still segmented by the original pkunlp tool, there would be a certain degree of robustness brought to the error correction results by different CWS systems. This may further cause a metric value drop for the error correction algorithm. Here is an example in Figure 1. For the standard answer "但是这种想法太短浅，而且有很大的错误。" (But this idea is too short-sighted and deeply flawed.), the result obtained by the LTP[2] tool is "但是 这种 想法 太 短浅，而且 有 很 大 的 错误 。" and the result obtained by the pkunlp tool is "但是 这 种 想法 太 短浅，而且 有 很 大 的 错误 。". Even though there are no grammatical errors or wrong correction actions in this sentence segmented by the LTP tool, the "这种" in the sentence is considered to be an incorrect correction action during the evaluation according to the segmentation by the pkunlp tool.

---

[1] http://59.108.48.37:9014/lcwm/pkunlp/downloads/libgrass-ui.tar.gz

**Table 1. An example of different tools.**

| Original Sentence | 但是这种想法太短浅，而且有很大的错误。 |
|---|---|
| Segmentation by pkunlp | 但是/这/种/想法/太/短浅/，/而且/有/很/大/的/错误/。 |
| Segmentation by LTP | 但是/这种/想法/太/短浅/，/而且/有/很/大/的/错误/。 |

(2) Since most of the existing CWS algorithms target normative text, the grammar errors in the text may lead to mis-segmentation if applying standard CWS algorithms to CGEC scenarios. In addition, the CWS algorithm itself suffers from ambiguous segmentations and unregistered words which may further introduce additional error information for CGEC.

(3) If researchers use different tools to segment the data when training their models, the results of their respective models are hard to be directly compared, e.g., the same model may produce different $F_{0.5}$ values when within different segmentation tools.

For the situations above, we argue that the evaluation metrics of the model in the CGEC task should not be influenced by the CWS tool, i.e., the evaluation metrics should be completely independent of the CWS algorithm. In addition to being influenced by the CWS algorithm, the current metrics for GEC are also influenced by language models. For example, grammaticality and fluency proposed by Asano et al. [1] require statistical language models to obtain evaluation scores. However, there are differences in the language models constructed from different data that different scores can be obtained for the same sentence with different language models. To this end, we propose three novel evaluation metrics for CGEC in two dimensions: reference-based and reference-less. On one hand, from the reference-based aspect, we propose two metrics, namely sentence-level accuracy and char-level BLEU, to evaluate the corrected sentences. Sentence-level accuracy is a hard accuracy, i.e., a valid correction sample needs to correct all the errors in the sentence correctly. To meet this requirement, we further extended the original CGEC test set of NLPCC 2018 so that each sentence could correspond to more answers. Besides, char-level BLEU is a soft evaluation metric that focuses on measuring the fluency and local accuracy of the corrected sentences. On the other hand, from the reference-less aspect, we introduce char-level meaning preservation to compare the difference between the corrected and original sentences. It is used to measure the semantic preservation degree of the corrected sentences. We deeply evaluate and analyze the reasonableness and validity of the three proposed metrics and expect them to become a new standard for CGEC.

In summary, the contributions of this paper are:

(1) To the best of our knowledge, we are the first to consider the negative impact of CWS algorithms on the CGEC evaluation.

(2) We extend the original test set of NLPCC2018 by producing more correction answers for each sentence.

(3) We propose three evaluation metrics applicable to CGEC from different dimensions.

## 2 Related Work

There are many evaluation metrics for GEC assessment, such as F-value [4], I-measure [5], $F_{0.5}$ [3] and GLEU (Generalized Language Evaluation Understanding) [8].

The HOO 2012 task [4] aims to identify and correct preposition and determiner errors in English, by calculating F-values from three perspectives: detection, recognition, and correction. I-measure [5] evaluates corrections at the token level using a globally optimal alignment between the source, a system hypothesis, and a reference. It provides scores for both detection and correction and is sensitive to different edit operations. Dahlmeier and Ng et al. [3] proposed a novel evaluation system MaxMatch to calculate the phrase-level sequence score $F_\beta$ between the source sentence and the corrected output (generally take β as 0.5). It is considered as one of the common evaluation metrics for GEC systems recently.

In the above approaches, the reference sentences may be difficult to cover all possible grammatical errors, resulting in punishment if some proper corrections are not included. This punishment mechanism is somewhat not reasonable. Asano et al. [1] introduced three reference-less metrics which cover syntax, fluency, and semantics evaluation to overcome the above limitations. Yoshimura et al. [11] further optimized the above three metrics (grammaticality, fluency, and meaning preservation) with three different BERT models. The enhanced metrics achieve the highest correlation with manual evaluation at both system and sentence levels. However, this method is not fine-grained enough and only three metrics are not sufficient to fully evaluate the corrected sentences. On the other hand, the $F_{0.5}$ and GLEU [8] metrics treat all grammatical errors equally, ignoring the ease of correcting different grammatical errors. To more comprehensively evaluate different GEC models, Takumi et al. [6] proposed a method to measure the difficulty of grammar correction where the more times a system corrects an error, the less difficult it is to correct the error. Recent difficulty evaluations and weighting algorithms can reflect the difficulty of grammar correction to some extent, which is useful for subsequent model effectiveness evaluation as well as correcting difficult grammatical errors.

Although there have been some studies on GEC evaluation metrics, these studies have largely focused on English which have not considered the influence of word segmentation for CGEC.

---

[2] https://github.com/HIT-SCIR/ltp

## 3 Assessments for CGEC

We propose three novel evaluation metrics for CGEC in two dimensions: reference-based and reference-less. In terms of the reference-based metric, we introduce sentence-level accuracy and char-level BLEU to evaluate the corrected sentences. Besides, in terms of the reference-less metric, we adopt char-level meaning preservation to measure the semantic preservation degree of the corrected sentences. Especially, suppose the original sentence is represented as $O_i$ and the sentence generated by the grammar error correction model is represented as $C_i$ while the gold-standard annotations are represented as $S_i = \{S_{i1}, S_{i2}, ..., S_{im}\}$ where $m$ indicates the number of gold-standard correction annotations for the $i$-th sentence.

### 3.1 Sentence-level accuracy

Sentence-level accuracy measures the accuracy of the model prediction results from the sentence level. In sentence-level accuracy measurement, if the corrected sentence $C_i$ completely matches one gold-standard annotation in $S_i$, it is considered as a valid correction:

$$y_i = \begin{cases} 1 & if\ C_i\ appears\ in\ S_i \\ 0 & if\ C_i\ not\ appears\ in\ S_i \end{cases} \quad (1)$$

where $y_i$ indicates whether $i$-th sentence is a valid correction. Then the sentence-level accuracy of the test set is:

$$Acc_{sen} = \frac{\sum_{i=0}^{N} y_i}{N} \quad (2)$$

where $N$ is the sample number of the test set.

### 3.2 Char-level BLEU

BLEU (Bilingual Evaluation Understudy) [9] evaluates the difference between the reference (gold-standard translation) and the candidate sentences generated by the model. In this paper, we extend BLEU to the CGEC task evaluation and propose a char-level BLEU. The modified $n$-gram ($n = 1,2,3,4$) precision for each order is calculated as follows:

$$P_n = \frac{\sum_i \sum_k \min(h_k(C_i), \max_{j \in m}(h_k(S_{ij})))}{\sum_i \sum_k \min(h_k(C_i))} \quad (3)$$

where $h_k(\cdot)$ stands for the number of the $k$-th $n$-gram appearing in the sentence and $\max_{j \in m}(h_k(S_{ij}))$ represents the highest frequency of a certain n-gram in $m$ gold-standard annotations.

To balance the effects of the modified $n$-gram precision, following Papineni et al., we take the logarithmic average of each modified $n$-gram precisions and weight it with $W_n$.

$$P'_n = \exp\left(\sum_{n=1}^{N} W_n \log(P_n)\right) \quad (4)$$

Among them, $W_n = \frac{1}{N}$ and $N$ refers to the phrase consisting of up to $N$ characters.

After computing the logarithmic average of the modified $n$-gram precisions $P'_n$, using $n$-grams up to length $N$ and positive weights $W_n$ summing to one, we further compute the brevity penalty (BP) to avoid the bias of the modified $n$-gram precisions $P'_n$. Specifically, let $l_c$ be the length of the corrected sentence $C_i$ and $l_s$ be the length of the gold-standard annotation which is closest to $l_c$, we compute the brevity penalty (BP) as follows:

$$BP = \begin{cases} 1 & if\ l_c > l_s \\ e^{1-\frac{l_s}{l_c}} & if\ l_c \leq l_s \end{cases} \quad (5)$$

Eventually, we can calculate the BLEU score as follows.

$$BLEU = BP \cdot P'_n \quad (6)$$

### 3.3 Char-level meaning preservation

Inspired by Hiroki et al. [1], we propose a reference-less metric, char-level meaning preservation. Instead of evaluating the similarity of revised sentences to gold-standard annotations, char-level meaning preservation computes the character-level similarity of the corrected sentence and the source sentence. Char-level meaning preservation assesses how much of the meaning is preserved between the original and corrected sentences. For the $i$-th sentence, the score $S_{CM}(C_i, O_i)$ for the original sentence $O_i$ and the corrected sentence $C_i$ is calculated as follows:

$$S_{CM}(C_i, O_i) = \frac{P \cdot R}{t \cdot P + (1-t) \cdot R} \quad (7)$$

$$P = \frac{m(C_i, O_i)}{|C_i|} \quad (8)$$

$$R = \frac{m(C_i, O_i)}{|O_i|} \quad (9)$$

where $m(C_i, O_i)$ denotes the matched character number between the corrected and original sentences, and $|C_i|$ and $|O_i|$ respectively represent the character number in the corrected and original sentences. Notably, we use $t = 0.85$, which is a default value provided by Hiroki et al.

When the char-level meaning preservation tends to 1, the similarity between the modified sentence and the original sentence is higher, that is, the model can obtain a high score without correcting errors at all, so the index is not as high as possible, but it is the optimal value when it approaches a certain value. Therefore, this indicator is further revised. The average value of the char-level meaning preservation of all answer sentences and the corresponding original sentences is calculated as $MP_{average}$, and the revised char-level meaning preservation is the absolute value of the difference between the original char-level meaning preservation and the average char-level meaning preservation $MP_{average}$:

$$MP' = |MP - MP_{average}| \quad (10)$$

When the value of the revised char-level meaning preservation is smaller, the content preservation degree of the sample is higher, and the modification amplitude is smaller.

**Table 2. Statistics of the datasets.**

|       | Source | Num. of sample |
|-------|--------|----------------|
| **Train** | NLPCC  | 1215690        |
|       | HSK    | 156870         |
| **Dev**   |        | 5000           |
| **Test**  |        | 2000           |

## 4 Dataset

The datasets used in this paper are the official Lang-8 Chinese dataset provided by NLPCC and the HSK [13] dataset. The Lang-8 dataset [7] is collected from the lang-8.com essay correction platform, where one error may have multiple corrections. The HSK dataset is a dynamic essay dataset built by Beijing Language and Culture University, collected from the answer scripts of the HSK essay examinations from 1992-2005. The development set consists of 5000 sentences from the training data provided by NLPCC, and the test set is chosen from the NLPCC 2018 open assessment competition. The data statistics are shown in Table 2.

As the sentence-level metric follows a strict evaluation criterion, the more answers for each sentence, the more objective assessment of the metric. We therefore re-annotate the test set provided by NLPCC 2018. The annotation process is comprised of two steps. (1) two annotators re-annotate sentences that originally contained only one standard answer, and ten annotators annotated sentences that originally had two answers. (2) the results generated from (1) are extended to the answer set if they occur more than twice, while for results that occur only once, they are then audited by other annotators. The extended new test set contains a total of 2885 revised samples, corresponding to a maximum of 10 answers for a sentence.

**Table 3. The results among different methods.**

|  | P | R | $F_{0.5}$ | $Acc_{sen}$ | $BLEU_c$ | $MP'$ |
|---|---|---|---|---|---|---|
| **CNN (word)** | 34.69% | 10.03% | 23.26% | 0.0715 | 0.8385 | 0.0360 |
| **CNN (word, simplified) with HSK** | 39.96% | 10.38% | 25.45% | 0.0820 | 0.8486 | 0.0385 |
| **LSTM (word)** | 31.48% | 11.61% | 23.45% | 0.0805 | 0.8381 | 0.0315 |
| **LSTM (word, simplified) with HSK** | 38.68% | 12.24% | 27.01% | 0.0935 | 0.8502 | 0.0354 |
| **Transformer (word)** | 34.04% | 11.47% | 24.42% | 0.0850 | 0.8391 | 0.0341 |
| **Transformer (char)** | 36.73% | 12.72% | 26.66% | 0.0880 | 0.8415 | 0.0342 |
| **Transformer (word) with HSK** | 39.51% | 13.20% | 28.25% | 0.0960 | 0.8469 | 0.0340 |
| **Transformer (char) with HSK** | 39.65% | 13.79% | 28.84% | 0.1095 | 0.8497 | 0.0355 |
| **Transformer (word, simplified) with HSK** | 41.18% | 13.17% | 28.90% | 0.0970 | 0.8494 | 0.0356 |
| **Transformer (char, simplified) with HSK** | 42.04% | 12.73% | 28.79% | 0.0935 | 0.8484 | 0.0366 |

## 5 Experiment

### 5.1 Experimental Settings

In this paper, we implement the CGEC model based on Fairseq[3] with PyTorch[4] as the backend. The source and target word embedding matrices share the same 512-dimensional word dictionary and model weights. The learning rate, dropout, epoch, and batch size in this paper are respectively set as 0.001, 0.2, 50, and 64. The optimal model is selected according to the loss of the development set. In the decoding stage, we set the beam search size to 5.

As for the evaluation metrics, we use $F_{0.5}$ to evaluate each model. Please note that since the official pkunlp[5] is no longer publicly available, our training and test sets are re-segmented with the LTP[6] tool. We also used OpenCC[7] to convert traditional Chinese characters in the corpus to simplified Chinese characters. We re-implement the current advanced baseline models with the re-segmented data and compare their relative metric values, which means that we cannot directly compare with existing paper methods. We evaluate and analyze the current state-of-the-art end-to-end models with the proposed metrics in this paper. The baseline models are as follows: CNN-based seq2seq model [2], LSTM-based seq2seq model [12] and Transformer [10].

### 5.2 Result

We evaluate the current state-of-the-art methods in CGEC with the proposed metrics as well as the commonly used metrics and the experimental results are shown in Table 3. In addition, we also conducted Pearson correlation tests between the three proposed metrics and the original $F_{0.5}$ metrics, and the results are shown in Table 4. $Acc_{sen}$ and $BLEU_c$ are highly correlated with $F_{0.5}$, among which the correlation between $Acc_{sen}$ and $F_{0.5}$ is the highest with a value of 0.8895, thus we use this metric as the first evaluation metric. Besides, $MP_c$ measures different objects from other metrics, and therefore the correlation between them is low. Additionally, we can see that the performance tends of different models are similar using $Acc_{sen}$ and $F_{0.5}$ as the evaluation metrics where the Transformer (word, simplified) with HSK and pretrained model performs best under both metrics. These results may show the reasonableness of our metrics.

---

[3] https://github.com/pytorch/fairseq
[4] https://github.com/pytorch/pytorch
[5] http://59.108.48.37:9014/lcwm/pkunlp/downloads/libgrass-ui.tar.gz
[6] https://github.com/HIT-SCIR/ltp
[7] https://github.com/BYVoid/OpenCC

**Table 4. The results about Pearson correlation tests.**

|  | $F_{0.5}$ | $Acc_{sen}$ | $BLEU_c$ | $MP'$ |
|---|---|---|---|---|
| $F_{0.5}$ | - | - | - | - |
| $Acc_{sen}$ | 0.8895 (0.0006) | - | - | - |
| $BLEU_c$ | 0.8321 (0.0028) | 0.7348 (0.0155) | - | - |
| $MP_c$ | 0.2733 (0.4449) | 0.0467 (0.8981) | 0.5741 (0.0826) | - |

## 5 Conclusion

In this paper, we observe that in the existing CGEC evaluation system, the evaluation values of the same error correction model can vary considerably under different word segmentation systems or different language models. This may lead to a lack of uniqueness and comparability in the evaluation of different methods. To this end, we study more objective evaluation metrics for Chinese Grammatical Error Correction (CGEC) and propose three novel evaluation metrics for CGEC. Sentence-level accuracy and char-level BLEU are reference-based metrics and char-level meaning preservation is a reference-less metric. In addition, to maintain the fairness of the sentence-level accuracy evaluation, we re-annotate and extend the test set provided by NLPCC 2018. We deeply evaluate and analyze the reasonableness and validity of the three proposed metrics as well as expect them to become a new standard for CGEC.